\documentclass{article}

\PassOptionsToPackage{numbers, compress}{natbib}

 \usepackage[preprint]{neurips_2026}

\usepackage{microtype}
\usepackage{graphicx}
\usepackage{subcaption}
\usepackage{booktabs} 

\usepackage{bbm}
\usepackage{amsmath, amssymb}
\usepackage{algorithm}
\usepackage{algorithmic}

\usepackage{graphicx}
\usepackage{multirow}
\usepackage{microtype}
\usepackage[dvipsnames, table, x11names]{xcolor}
\usepackage{colortbl}
\usepackage[utf8]{inputenc}
\usepackage[T1]{fontenc}

\usepackage{enumitem}

\usepackage{array, booktabs, boldline} %
\usepackage{cellspace}
\usepackage{hyperref}

\usepackage{amsmath}
\usepackage{amssymb}
\usepackage{mathtools}
\usepackage{amsthm}

\usepackage{wrapfig}

\usepackage{tikz}
\usetikzlibrary{arrows.meta} 
\usepackage{tcolorbox}

\usepackage{amsmath,amsfonts,bm}

\def\eqref#1{eq.~\ref{#1}}

\def\1{\bm{1}}

\def\vomega{{\bm{\omega}}}

\def\vb{{\bm{b}}}

\def\vk{{\bm{k}}}

\def\vo{{\bm{o}}}
\def\vp{{\bm{p}}}
\def\vq{{\bm{q}}}

\def\vv{{\bm{v}}}

\def\vx{{\bm{x}}}

\def\mI{{\bm{I}}}

\def\mO{{\bm{O}}}

\def\mR{{\bm{R}}}

\def\mW{{\bm{W}}}

\DeclareMathAlphabet{\mathsfit}{\encodingdefault}{\sfdefault}{m}{sl}
\SetMathAlphabet{\mathsfit}{bold}{\encodingdefault}{\sfdefault}{bx}{n}

\def\sE{{\mathbb{E}}}

\def\sT{{\mathbb{T}}}

\usepackage{amsthm}
\usepackage{amssymb}

\theoremstyle{definition}

\theoremstyle{remark}

\newcommand*{\prob}[1]{\mathbb{P}}

\newcommand{\R}{\mathbb{R}}

\def\vbeta{{\boldsymbol{\beta}}}

\usepackage{empheq}
\usepackage{fancybox}
\usepackage{tcolorbox}

\theoremstyle{plain}

\usepackage{listings}
\tcbuselibrary{listings,skins,breakable}

\lstdefinestyle{promptstyle}{
  basicstyle=\ttfamily\small,
  breaklines=true,
  breakatwhitespace=false,
  columns=fullflexible,
  keepspaces=true,
  showstringspaces=false,
  upquote=true,
  tabsize=2,
  literate=
    {’}{{'}}1
    {‘}{{'}}1
    {“}{{``}}1
    {”}{{''}}1
    {–}{{--}}1
    {—}{{---}}1
    {…}{{\ldots}}1
}

\newtcblisting{promptbox}[2][]{%
  listing engine=listings,
  listing only,
  listing options={style=promptstyle},
  colback=black!2,
  colframe=MidnightBlue!60!black,
  coltitle=white,
  coltext=black,
  fonttitle=\bfseries,
  title={#2 prompt},
  breakable,
  sharp corners,
  boxrule=0.8pt,
  left=1mm, right=1mm, top=1mm, bottom=1mm,
  enhanced,
  borderline west={2pt}{0pt}{MidnightBlue},
  attach boxed title to top left={yshift=-2mm, xshift=2mm},
  boxed title style={
    colback=MidnightBlue!80!black,
    boxrule=0pt,
    sharp corners,
    interior style={opacity=0.95}
  },
  #1 
}

\usepackage[disable,textsize=tiny]{todonotes}

\newcommand{\lanerope}{LaneRoPE}

\title{LaneRoPE: Positional Encoding for Collaborative Parallel Reasoning and Generation}

\author{
Gabriele Cesa \\
Qualcomm AI Research\thanks{
Qualcomm AI Research is an initiative of Qualcomm
Technologies, Inc.\\
\copyright{}2026 Qualcomm Technologies, Inc. and/or its affiliated
companies. All Rights Reserved. \\
Correspondence to: \texttt{gcesa@qti.qualcomm.com}
} \\
\And Thomas Hehn \\  Qualcomm AI Research$^*$ \\
\And Aleix Torres-Camps  \\  Qualcomm AI Research$^*$ \\
\And \`{A}lex Batlle Casellas \\  Qualcomm AI Research$^*$ \\
\And Jordi Ros-Giralt \\  Qualcomm AI Research$^*$ \\
\And Arash Behboodi \\  Qualcomm AI Research$^*$ \\
\And Tribhuvanesh Orekondy  \\  Qualcomm AI Research$^*$ \\
}

\begin{document}

\maketitle

\begin{abstract}
Parallel LLM test-time scaling techniques (e.g., best-of-$N$) require drawing $N>1$  sequences conditioned on the same input prompt.
These methods boost accuracy while exploiting the computational efficiency of batching $N$ generations.
However, each sequence in the batch is traditionally generated independently and hence does not reuse intermediate generations, computations, or observations from other sequences.
In this paper, we propose \lanerope{} to enable coordination and collaboration among $N>1$ sequences at generation time.
\lanerope{} involves two key ideas:
(a) an inter-sequence attention mask to make sampling of sequences dependent on one another; and
(b) a RoPE extension that injects positional information that captures relative positions between tokens, both within and outside a particular sequence.
We evaluate our approach on mathematical reasoning tasks and find promising results: 
\lanerope{} enables collaboration among sequences, yielding additional accuracy gains under limited generated sequence length.
Importantly, since \lanerope{} enables coordination with minimal changes to the underlying LLM architecture and introduces a negligible overhead at inference time, it is appealing to rapidly incorporate parallel reasoning into existing LLM inference pipelines.
\end{abstract}

\section{Introduction}

Parallel test-time scaling methods improve LLM performance with additional inference-time compute instead of finetuning \citep{cobbe2021training,brown2024large,wu2025inference,snell2024scaling,wang2022self}.
The key idea of these methods is generating \textit{multiple} responses (each potentially with its own reasoning trace) for a given input, and optionally, ranking and aggregating responses with a reward model.
The multiple responses are typically generated in parallel during the same forward pass (i.e., batched inference), and benefit from a hardware efficiency standpoint, since this amortizes memory transfers, minimizes kernel launch overheads, and better utilizes compute cores at each generation step.
Despite its efficiency, parallel inference is largely \textit{uncoordinated}: for a given input, the output sequences are generated independently.
This prohibits exploiting inherent decomposable problem structures \citep{yang2025multiverse}, limits the sharing of intermediate outputs between sequences, and the generation of diverse responses.

\begin{figure*}[t]
  \begin{center}
    \centerline{\includegraphics[width=\linewidth]{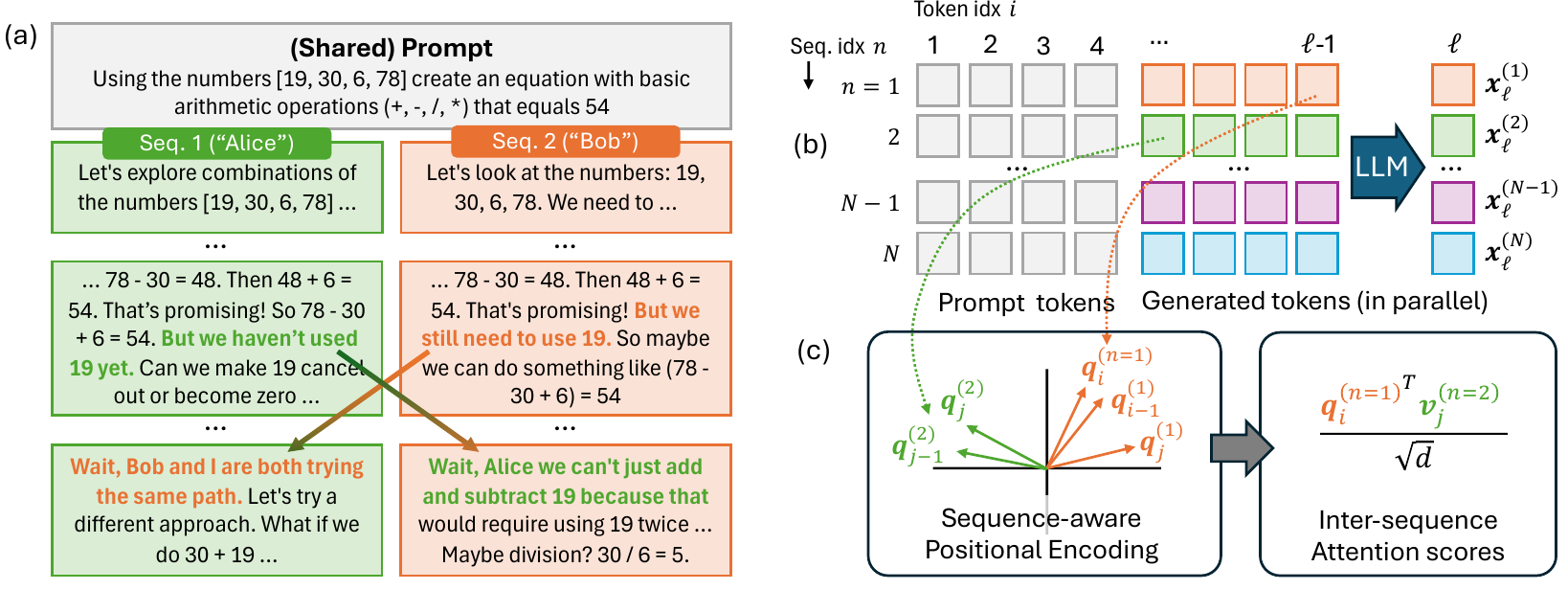}}
    \caption{
      \textbf{LaneRoPE: High-level introduction and intuition.}
      (a) We investigate the problem of \textit{collaborative reasoning}, where given a single input prompt, parallel sequences can reason by conditionally attending to other sequences mid-generation;
      (b) Our work assumes the tokens across sequences are generated in parallel using the same model, and thereby benefit from \textit{batched efficiency};
      (c) A key contribution of our work is introducing \textit{cross-lane attention}, such that attentions can be calculated across sequences (i.e., \emph{lanes}). To enable this, we introduce a novel position encoding scheme to account for relative distances between tokens of different sequences.
    }
    \label{fig:intro}
  \end{center}
\end{figure*}

Building on powerful parallel inference, how can we make multiple sequences coordinate with each other during generation?
Two families of approaches have gained traction, each with their inherent trade-offs.
The first family of approaches leverage `explicit branching' \citep{gandhi2024stream,pan2025apr,yang2025multiverse,zheng2025parallelr1} mechanisms: at inference time, a single sequence adaptively fans-out to multiple search paths (e.g., one path per decomposed subproblem) and funnels them into a single response.
Although they encourage the LLM to explicitly decompose tasks, the dynamic nature of how variable number of parallel processes are spawned and merged introduces challenges in predicting runtime compute resources and more importantly, they depend on specialized run-time engines.
The second family of `collaborative' approaches \citep{hsu2025groupthink,rodionov2025hogwild,dong2025gps}, under which our approach falls, introduce a more flexible means of coordination primarily by enabling shared visibility (Fig. \ref{fig:intro}a) into intermediate generations of all parallel sequences.
These techniques are practically attractive, since they rely on a fixed number of generation sequences (i.e., constant batch size; see Fig. \ref{fig:intro}b) and can hence largely leverage batching support by both LLM inference engines and the underlying hardware.
In particular, existing techniques \citep{hsu2025groupthink,rodionov2025hogwild} exploit an interesting insight that multiple sequences can be `virtually ordered' into a single sequence by reparameterizing the positions or positional encodings.
However, we argue that this is inherently a multi-sequence generation problem and inference must accurately reflect positions both along the token dimension (as typically done) \textit{and} sequence (or \emph{lane}) dimension.
In the rest of this manuscript, we use \emph{lane} and \emph{sequence} interchangeably to refer to batch elements generated simultaneously and inter-dependently. 

In this work, we propose \lanerope{} to better enable parallel reasoning and generation.
Similar to parallel test-time scaling techniques \citep{wang2022self,wu2025inference} and collaborative reasoning \citep{hsu2025groupthink,rodionov2025hogwild,dong2025gps}, we assume a constant number of sequences (Fig. \ref{fig:intro}b) generated at inference.
Our approach involves two key insights.
First, we make intermediate reasoning traces and outputs visible across sequences by conditioning the next token on previous tokens from \textit{all sequences}.
We achieve this by using causal cross-sequence attention masks and generalizing attention score calculations to extend to inter-sequence tokens.
Second, we extend rotary positional encoding `RoPE' \citep{su2021roformer} to additionally capture relative distances among sequences (Fig. \ref{fig:intro}b), allowing us to learn the attention structure across parallel generations in a dynamic and task-dependent manner.
Our formulation is expressive enough to capture a range of parallel inference techniques, ranging from independent sampling (e.g., self-consistency) to the fully dense attention masks used in GroupThink \citep{hsu2025groupthink}.

Our evaluation of \lanerope{} on standard mathematical reasoning datasets (e.g., AIME, MATH500) and two open-weight LLMs shows promising results.
Overall, we find that the LLMs demonstrate a reliable accuracy improvement across multiple benchmarks. 
The flexibility in our formulation enables \lanerope{} to be integrated into existing models with minimal modification and, importantly, in contrast to prior work \citep{hsu2025groupthink,rodionov2025hogwild}, admits finetuning.
Furthermore, finetuning requires very few additional parameters ($<0.5\%$ over base models), which can be fit with simple SFT and synthetic data.
In summary, we find \lanerope{} takes a promising step towards enabling parallel \textit{collaborative} reasoning by fundamentally exploiting relative positions of tokens, both within and outside a particular sequence.

\paragraph{Contributions}
We summarize our contributions as:
\begin{itemize}[noitemsep,topsep=0pt,parsep=0pt,partopsep=0pt]
    \item A novel method to enable \emph{fine-grained token-level collaboration} during batched parallel generation of language models. Notably, our method encompasses some existing techniques but its superior flexibility and generality allow for further fine-tuning.
    \item A recipe to synthesize \emph{collaborative reasoning traces} which can be used for supervised fine-tuning (SFT).
    \item A fine-tuning pipeline to unlock and strengthen the parallel reasoning capabilities of pre-trained models.
\end{itemize}

\section{Related Work}
\label{sec:relatedworks}

\paragraph{LLMs, Reasoning and Test-time Search. }
Large language models (LLMs) have attracted significant attention, largely due to their remarkable performance across a range of tasks.
Many of these tasks (e.g., math, coding) test the LLM's ability for generalization, abstraction, and multi-step problem solving.
Especially for such tasks, reasoning \citep{wei2022chain,kojima2022large} has proven to be a powerful paradigm and has significantly improved success rates at solving challenging problems.
These techniques can be further extended to explore various reasoning paths before concluding with a final solution.
The reasoning search paths can take various structures, but can broadly be classified into sequential \citep{muennighoff2025s1}, graph-like \citep{yao2023tree,besta2024graph}, or parallel \citep{wang2022self,brown2024large}.
In this work, we primarily study reasoning as a parallel search problem.

\paragraph{Parallel Reasoning: Independent and Inter-dependent. }
A popular strategy to scale compute at test-time to improve accuracies is with parallel generations \citep{wang2022self,cobbe2021training}.
For a given problem, multiple sequences (potentially chain-of-thought reasoning followed by a solution) from the same LLM are sampled in parallel and ranked (e.g., using a reward model) to determine the final solution.
Prior works \citep{wang2022self,cobbe2021training,brown2024large,snell2024scaling,wu2025inference} primarily demonstrate effectiveness where the sequences are sampled \textit{independently}.
This however leads to inefficient use of compute, since there is little collaboration between sequences and risks redundant compute between sequences.
A new recent line of work explores whether the sequences can be sampled \textit{inter-dependently} \citep{dong2025gps}, such that LLMs can decide to adaptively parallelize and coordinate \textit{during} generation.
This line of inter-dependent generation broadly falls under two categories:
(a) `adaptive branching' \citep{jin2025pasta,pan2025apr,qi2025ssa,yang2025multiverse,chen2025aspd,zheng2025parallelr1,wen2025parathinker,lian2025threadweaver} where at generation time, special tokens (e.g., \texttt{fork} and \texttt{merge}) signal the inference engine to spawn or merge multiple parallel sequence generations; and
(b) `fine-grained collaboration', as in \cite{hsu2025groupthink,rodionov2025hogwild} and recently \cite{dong2025gps}, where sequence generation is always parallel (constant batch size) and each sequence can attend to part or all of the tokens generated by other sequences using dynamic attention masks.
Our approach falls into the second category, and unlike previous works we introduce a flexible framework to positionally encode cross-sequence tokens which additionally account for relative inter-sequence distances.

\paragraph{Positional Encoding in LLMs}
Large language models (LLMs) incorporate \textit{positional encodings} to inject sequence order information into transformer architectures, since the self-attention mechanism alone is permutation-invariant \citep{vaswani2017attention}. 
The original Transformer used fixed sinusoidal position encodings \citep{vaswani2017attention}, while subsequent models like BERT adopted learned positional embeddings \citep{devlin2019bert}, both providing each token with an absolute position bias. 
To improve generalization beyond fixed-length contexts, \textit{relative positional encodings} were introduced to encode pairwise token distances directly in self-attention \citep{shaw2018self}, allowing the model to focus on relative order. 
More recent approaches modify how positional information enters the attention mechanism to enable better extrapolation to longer sequences. For example, \emph{rotary positional encodings} (RoPE) rotate query/key vectors by angle proportional to token index, effectively combining absolute and relative positional information \citep{su2021roformer}. Similarly, \emph{ALiBi} (Attention with Linear Biases) eliminates explicit embeddings and instead adds a distance-dependent bias to attention scores, allowing models trained on shorter sequences to generalize to much longer inputs \citep{press2022alibi}. 
In our work, we extend RoPE \citep{su2021roformer} to additionally model relative distances between sequences.

\section{\lanerope{}: Cross-lane positional encoding}
\label{sec:method}

In this section, we present our approach \lanerope{} to enable collaborative parallel inference.
We begin by covering preliminaries, and then we detail our approach.

\begin{figure*}[t]
  \begin{center}
    \centerline{\includegraphics[width=\linewidth]{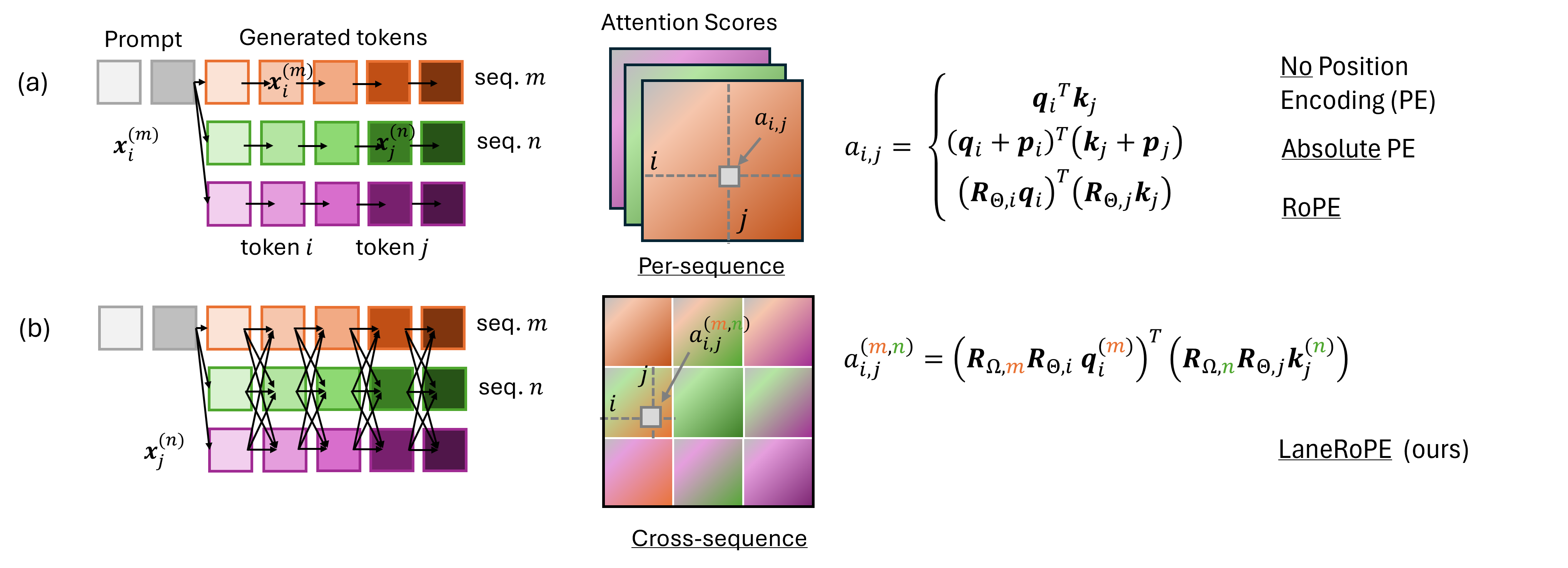}}
    \caption{
      Comparison of RoPE \citep{su2021roformer} and \lanerope{} for parallel inference.
      (a) Parallel inference (e.g., best-of-N) relies on generating sequences independently. As a result, positional encodings and attention scores are defined only per-sequence.
      (b) With \lanerope{}, tokens are generated by causally attending to tokens from all sequences. We achieve this by introducing cross-sequence attention scores with a novel position encoding scheme to account for inter-sequence relative distances.
    }
    \label{fig:method}
  \end{center}
\end{figure*}

\subsection{Preliminaries}

A standard transformer self-attention layer \citep{vaswani2017attention} takes token embeddings $\{\vx_i\}_{i=1}^{L}$, computes query, key, value representations (Eq. \ref{eq:qkv}) and produces corresponding attention logits and outputs (Eq. \ref{eq:self-attn}):
\begin{align}
    \vq_i = f_q(\vx_i, i) ,\quad \vk_j = f_k(\vx_j, j) ,\quad \vv_j = f_v(\vx_j, j) \label{eq:qkv} \\
    \alpha'_{i, j} = \text{softmax}_j\!\left(\vq^{\top}_i \vk_j/\sqrt{d}\right) ,\quad \vo_i = \sum_{j=1}^{L} \alpha'_{i, j} \vv_j \label{eq:self-attn}
\end{align}

\paragraph{Positional Encodings and RoPE}
Positional encodings injected into $f_{\{q,k\}}$ make outputs position-dependent.
RoPE \citep{su2021roformer} enforces that attention logits depend only on the relative offset $i-j$ by rotating embeddings with a block-diagonal matrix $\mR_{\Theta,i}$ (see Appendix~\ref{apx:preliminaries}):
\begin{align}
f_{\{q,k\}}(\vx_i, i) &= \mR_{\Theta,i}\, (\mW_{\{q,k\}} \vx_i + \vb_{\{q,k\}}) \\
\mR_{\Theta,i} &= \mathrm{diag}\!\left(R(i\theta_1),\, \ldots,\, R(i\theta_{d/2})\right)  && \text{with frequencies params.:  } \Theta = (\theta_l)_{l=1}^{d/2} \label{eq:rope-rot-mat} \\
R(i\theta_\ell) &=
\begin{bmatrix}
\cos(i\theta_\ell) & \sin(i\theta_\ell) \\
-\sin(i\theta_\ell) & \cos(i\theta_\ell)
\end{bmatrix}  \in \mathbb{R}^{2\times 2} && \ell = 1,\ldots,d/2
\end{align}

\paragraph{Parallel Test-time Scaling}
Parallel test-time scaling \citep{brown2024large,snell2024scaling,wu2025inference} improves LLM accuracy at inference time by sampling $N>1$ sequences and selecting the best via majority vote \citep{wang2022self} or a verifier \citep{cobbe2021training,lightman2024lets}.
Most prior work assumes \textit{independent} sampling across sequences.

\subsection{\lanerope{}: Sequence-aware Positional Encoding}

\paragraph{Multi-sequence generation}
We generate $N>1$ sequences $\{\vx^{(n)}\}_{n=1}^N$ auto-regressively in parallel for the same prompt:
\begin{align}
    p(\vx^{(n)}_{i+1} &\; | \; \texttt{prompt},\; \{\vx^{(m)}_{1:i}\}_{m=1}^N) \label{eq:gen-batched}
\end{align}
Here, as opposed to \textit{independent} sampling (conditional generation of $\vx^{(n)}_{i+1}$ on $\vx^{(n)}_{1:i}$), we now have \textit{inter-dependent} generation (conditioned on $\{\vx^{(m)}_{1:i}\}_{m=1}^N$) via \emph{causal inter-sequence attention}, allowing tokens in sequence $m$ to attend to tokens in sequence $n$:
\begin{align}
    {\alpha'}_{i, j}^{(m, n)} &= \text{softmax}_{j,n}\!\left({\vq_i^{(m)}}^{\top} \vk_j^{(n)}/\sqrt{d}\right) \label{eq:self-attn-2d}
\end{align}
However, since tokens at the same time step across sequences share identical positional indices, standard RoPE cannot distinguish their sequence identities.

\paragraph{Cross-sequence Positional Encoding}
To distinguish between sequence identities, we introduce a second block-diagonal rotation $\mR_{\Omega,m}$ that encodes the sequence index $m$:
\begin{align}
    \vq_{i}^{(m)} &= f_q(\vx_{i}^{(m)}, i, m) ,\quad \vk_{j}^{(n)} = f_k(\vx_{j}^{(n)}, j, n) \label{eq:qkv-2d} \\
    f_{\{q, k\}}(\vx_i, i, m) &= \mR_{\Omega, m} \mR_{\Theta, i} (\mW_{\{q,k\}} \vx_i^{(m)} +\vb_{\{q,k\}})  \label{eq:lanerop_proj}
\end{align}
with lane frequencies $\Omega = \{\omega_t\}_{t=1}^{d/2}$.
The associativity $R(\omega_t m) R(\theta_t i) = R(\omega_t m + \theta_t i)$ lets both rotations be merged and applied together, enabling drop-in integration into any RoPE architecture.
Analogously, while RoPE uses a 1D Fourier basis over the token sequence, \lanerope{} uses a 2D Fourier basis over the joint (token, sequence) grid.
An important effect of $\mR(\Omega_t, m)$ being orthogonal is that the (pre-softmax) attention weight between two tokens $i, j$ within the same sequence $m$ is unchanged since both key and query vectors are rotated by $\mR(\Omega_t, m)$ and, therefore, the inner product ${\vq_i^{(m)}}^{\top} \vk_j^{(m)}$ remains the same.
In particular, we emphasize that for $N=1$ lane, \lanerope{} simplifies to standard RoPE.

\subsection{\lanerope{}: Properties and Initialization Strategies}
\label{sec:initialization_strategies}
Thanks to its flexibility, \lanerope{} can express some existing parallelization strategies, but can also go beyond if trained.
This suggests using one (or a combination) of these strategies to \emph{initialize \lanerope{} from pre-trained models}, in order to leverage their native reasoning and cooperation capabilities.

\paragraph{Special Case: GroupThink}
GroupThink \citep{hsu2025groupthink} previously proposed causal inter-sequence attention by `virtually' ordering generation of multiple sequences, albeit within a single sequence.
We find this to be a special case of our proposed approach and elaborate further.
GroupThink distinguishes sequences by assigning token $\vx_i^{(m)}$ the virtual position $Km+i$ (sequence gap $K$), so that within-sequence tokens appear close in RoPE-space while cross-sequence tokens appear further away.
This is exactly recovered in \lanerope{} via the rotation-matrix decomposition
\begin{align}
   \mR_{\Theta,Km+i} = \mR_{\Theta,Km}\mR_{\Theta,i} = \mR_{K\Theta,m}\mR_{\Theta,i} = \mR_{\Omega,m}\mR_{\Theta,i}
\end{align}
with $\Omega := \{K\theta_t\}$, i.e.\ GroupThink is \lanerope{} with lane frequencies $\Omega = K\Theta$.
A \emph{limitation} is that tokens from higher-index sequences ($n>m$) appear at \emph{negative} virtual relative positions when attended by sequence $m$, which is out-of-distribution for causally pre-trained models.\footnote{Hogwild! \citep{rodionov2025hogwild} avoids this by dynamically reordering sequences, but requires custom attention kernels.}

\begin{figure}[t]
  \centering
  \begin{subfigure}{0.44\linewidth}
    \centering
    \includegraphics[width=\linewidth]{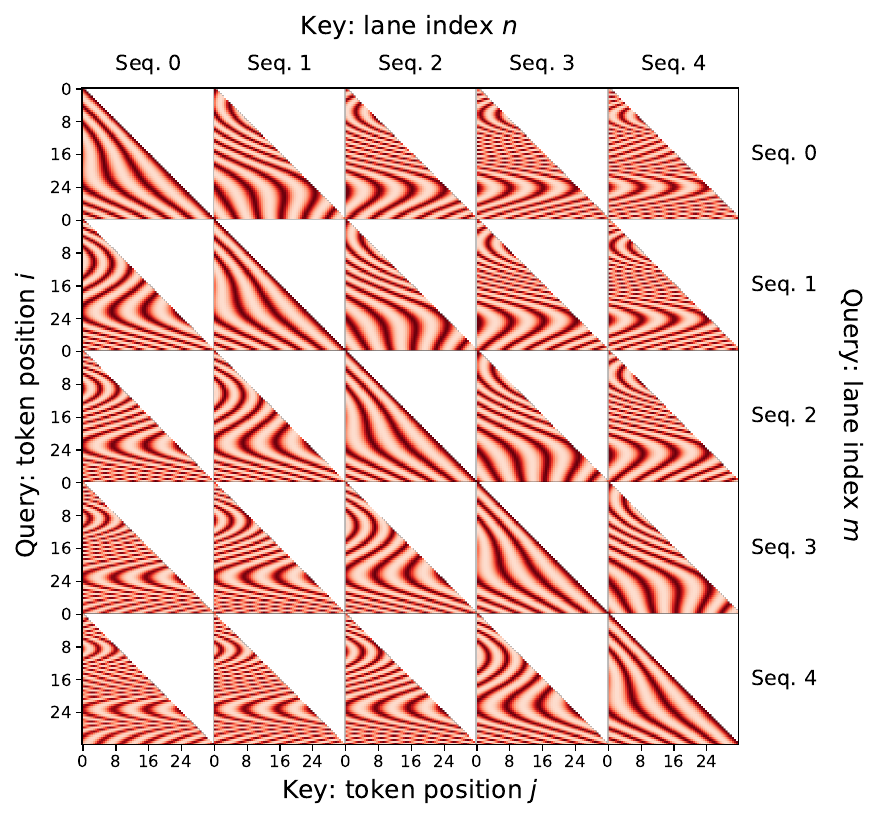}
    \caption{$\beta(0) = 0$, i.e. $|\hat{\vbeta}|_2 = 0$}
    \label{fig:four-a}
  \end{subfigure}\hfill
  \begin{subfigure}{0.44\linewidth}
    \centering
    \includegraphics[width=\linewidth]{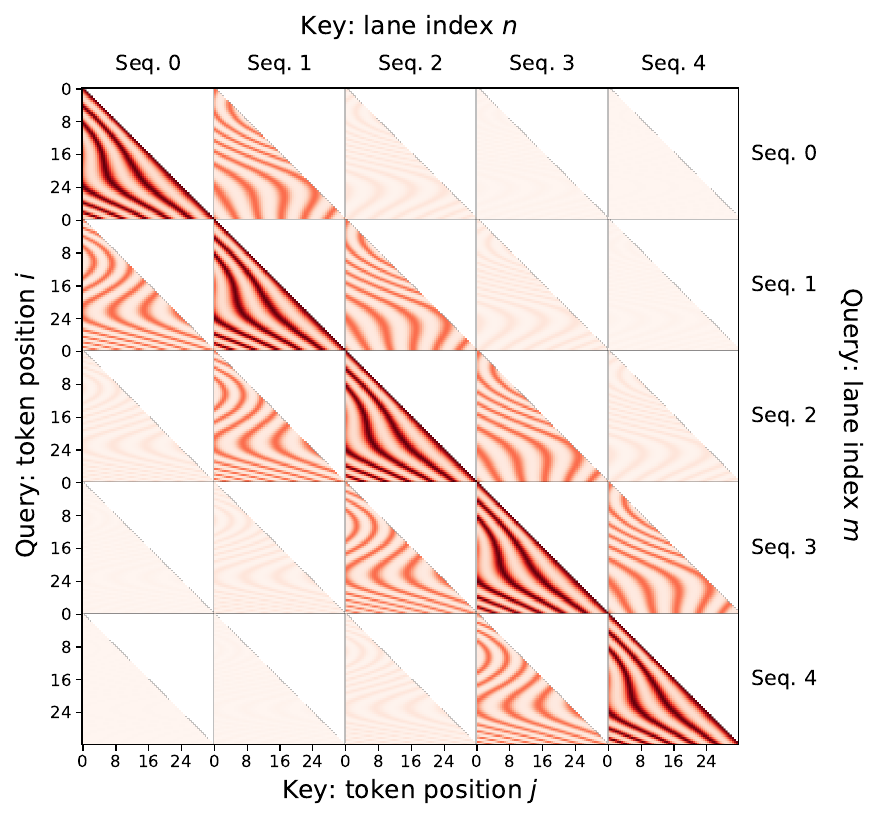}
    \caption{$\beta(0) > 0$, i.e. $|\hat{\vbeta}|_2 > 0$}
    \label{fig:four-c}
  \end{subfigure}
  \caption{Effect of the attention bias $\beta$ on attention scores in \lanerope{}, controlled by $|\hat{\vbeta}|_2$. A GroupThink initialization is used, where tokens from different lanes occupy distinct virtual position indices. The causal mask within each lane is preserved.}
  \label{fig:lanerope_attention_pattern_beta}
\end{figure}

\paragraph{Special Case: Parallel Independent Sampling}
Independent sampling across $N$ sequences is equivalent to adding a sequence-identity bias $\beta(m-n) \propto \delta_{[m=n]}$ to the attention logits, so that cross-lane scores are suppressed:
\begin{align}
    \label{eq:attention_bias}
    a_{i, j}^{(m, n)} &= {\vq_i^{(m)}}^{\!\top} \vk_j^{(n)} + \beta(m-n)
\end{align}
We approximate $\beta$ via a $F$-dimensional discrete Fourier basis and incorporate it into \lanerope{} by augmenting the projection layers with $F$ additional bias dimensions (see Appendix~\ref{apx:initializationstrategies}):
\begin{align}
\label{eq:lanerope_independent_init}
    \mW_{\{q,k\}}' := \begin{pmatrix} \mW_{\{q,k\}} \\ \mO \end{pmatrix} \ ,\quad
    \vb_{\{q,k\}}' := \begin{pmatrix} \vb_{\{q,k\}} \\ \hat{\vbeta} \end{pmatrix}
\end{align}
with $\mO \in \mathbb{R}^{F\times d_\text{emb}}$ zero-initialized.
This leaves all pre-trained weights intact and adds negligible overhead 
(${\approx}0.25$B parameters for $F{=}2$ in \texttt{Qwen2.5-7B}).
The bias strength is controlled by $|\hat{\vbeta}|_2$; Fig.~\ref{fig:lanerope_attention_pattern_beta} shows its effect on the resulting attention pattern.

\paragraph{Initialization Strategy: NTK-aware correction for GroupThink}
To mitigate the \emph{negative virtual indexes limitation} of GroupThink, we draw inspiration from the "NTK-aware" interpolation methods like YaRN~\cite{pengyarn} which modify RoPE's frequencies to extend the context length of pre-trained models.
The main idea is that, despite the translation equivariance of RoPE, frequencies with a relatively long period compared to the pre-trained context length enable the model to encode absolute positions, rather than relative ones.
We suspect these low frequencies suffer the most from the negative relative indexes of GroupThink, which fall outside the portion of the period typically observed during pre-training.
Hence, we introduce a scaling function $\gamma(\theta_t) \in [0, 1]$ which smoothly decays to $0$ the frequencies below a certain threshold, i.e. $\Omega := \{\gamma(\theta_t) K \theta_t\}_t$, such that the lane index has limited effects on them and they maintain their capacity to represent absolute positions; see Appendix~\ref{apx:initializationstrategies} for more details.

In practice, we combine these strategies: lane frequencies $\vomega' = K\Theta$ (GroupThink or NTK) for cross-lane structure, and the Fourier bias for initialization to independent sampling.
Since $\mW_{\{q,k\}}'$ and $\vb_{\{q,k\}}'$ are fully learnable, training is free to move beyond these special cases.
\section{Training \lanerope{}}
\label{sec:training}

To elicit cross-lane interactions in a base model, we consider two fine-tuning strategies: \emph{i)} supervised finetuning (SFT) on a dataset that mimics collaborations, and \emph{ii)} optimizing with a training objective adapted from Kahneman-Tversky Optimization (KTO) \cite{ethayarajh2024kto} on a dataset of (un)successful reasoning traces independently generated by the base model.

\subsection{SFT with collaborative data}
We construct this dataset using \texttt{Qwen3-30B-A3B-Thinking-2507}~\citep{qwen3technicalreport}, using questions from the \emph{DeepScaleR-Preview-Dataset}~\citep{luo2025deepscalermodel}.
To simulate collaboration across $N$ lanes using a sequential LLM, we prompt the LLM to play the role of $N$ different assistants in a step-by-step manner.
An assistant's message at each step ends after $128$ new tokens.
After each step, an assistant is prompted to generate a new message by completing its own answer, pre-filled with the text it generated so far, which follows a user message providing the query and the text previously generated by other assistants.
For the final SFT dataset, we separate the text generated by each assistant into individual traces such that each trace only contains the messages of one assistant.
As a result, a model fine-tuned on this dataset requires a form of cross-sequence interaction to be able to fit the data.
In total, we generate $3$ conversations with $N=2, 3$ or $4$ assistants on a subset of questions of the dataset, resulting in about $50\mathrm{k}$ raw conversations.
See Apx.~\ref{apx:synthetize} for more details.
To focus on quality over quantity, we further filter the original generations based on the following criteria.
First, we only keep samples where all the assistants give the correct answer.
Second, we employ a length limit in range $10-30$ messages per lane and remove all samples outside that range.
Third, we use a simple heuristic to filter out conversations with little or no interactions between the assistants.
After this curation process, the final SFT dataset consists of $4797$ conversations.
See Apx.~\ref{apx:sft_examples} for an example of the SFT dataset.

\subsection{KTO with independent reasoning traces}

We sample $8$ completions with the base model (either \texttt{DeepSeek-R1-Distill-Qwen-1.5B} or \texttt{DeepSeek-R1-Distill-Qwen-7B}) and a maximum completion length of $4096$, for each of the $40315$ questions of the \emph{DeepScaleR-Preview-Dataset}~\citep{luo2025deepscalermodel}. 
Among these, we keep only those queries where \emph{at most half} of the answers were correct, to ensure a sufficiently challenging dataset.
Importantly, we preserve both good and wrong completions to be used for the KTO loss \cite{ethayarajh2024kto}, which respectively maximizes and minimizes their likelihoods, up to some per-sample saturation.
Hence, we create a dataset by grouping the completions of each query into subsets of $N=2, 3, 4$ lanes, ensuring each subset contains at least one correct answer, for about $15$k total groups; during training, \lanerope{} is able to cross-attend among the $N$ lanes within each subset.
While SFT with independently sampled data would force our model to mimic the i.i.d. distribution, batches with mixed good/wrong completions can \emph{encourage a non-trivial joint distribution among parallel lanes via the KTO loss}: e.g. it's easier to minimize the likelihood of a wrong completion when its final answer doesn't match that of the other completions in the batch.
See Appendix~\ref{apx:experiment_details}.

\subsection{Other training details}
To train our models, we first initialize them from open-source pre-trained LLMs using a combination of the initialization strategies described earlier.
In particular, we use a "sequence gap" $K=8192$ to bias the model towards a GroupThink-like behavior but also set a strong positional bias ($|\hat{\vbeta}|_2 \geq 1000.$) to encourage the model to attend to tokens in other sequences only when necessary.
Then, we train our models for one epoch on these curated datasets.
In our experiments with small $1.5B$, we train all their weights while we use LoRA (including trainable biases) for the $7B$ ones to reduce memory cost at training time and fit at least $4$ lanes within a single GPU.
When indicated with a $*$, we also train the frequencies $\Omega$ of the \lanerope{} itself.

\section{Experiments}
\label{sec:experiments}
To assess the effectiveness of our proposed method, we conduct evaluations across several reasoning benchmarks and models that measure diverse problem-solving competencies. 
We also verify the efficiency of \lanerope{} in Sec.~\ref{apx:inference_time}, where we see that its cross-lanes attention comes with negligible overhead.

\subsection{Setup}
\label{sec:experiments_setup}

\paragraph{Datasets}
Specifically, we benchmark performance on the MATH500 \citep{lightman2024lets}, the AMC 23 \citep{amc2023}, and AIME 24 and 25 \citep{aime2024,aime2025} datasets, all of which consist of challenging mathematical problems.

\paragraph{Models and Baselines}
We test our method on the fine-tuned reasoning models \texttt{DeepSeek-R1-Distill-Qwen} with 1.5B and 7B parameters \citep{deepseekai2025deepseekr1incentivizingreasoningcapability}.
To compare to existing work, we utilize the official implementation of Hogwild! \citep{rodionov2025hogwild} and evaluate it on the same datasets and models as \lanerope{}.
Due to lack of public code and to avoid potential differences in hyperparameters and evaluation implementation (see also Appendix~\ref{apx:accuracy}), we further implement a Bridge~\cite{dong2025gps} baseline ourselves, which we train and evaluate like our models.
In our experiments, each multi-lane model is tested over different $N\geq1$ parallel sequences.
Moreover, we sample $M$ sequences for each query in each benchmark: these are sampled independently in standard baselines but, when using $N$-lanes (e.g. $N{=}2$) models, we independently sample $M/N$ (e.g. $8$) groups of $N$ inter-dependent completions.
Finally, all generations use a temperature of $0.6$ and \texttt{top-p} value of $0.95$ (as recommended in \cite{deepseekai2025deepseekr1incentivizingreasoningcapability}) and are limited by $4096$ maximum tokens.
All training runs were conducted on Nvidia A100 GPUs, with a combination of single- and multi-GPU training runs on single nodes.

\begin{table}[h]
    \centering
    \begin{tabular}{Sl  c  c c c c c c }
    \textbf{Model}                        & \textbf{Train.}      & \textbf{N}     & \textbf{MATH500} & \textbf{AIME24} & \textbf{AIME25} & \textbf{AMC23} & \textbf{Avg.}           \\
        \hline
    DS-Qwen-1.5B                          &                      & 1              &           $79.2$ &          $19.4$ &          $19.1$ &         $56.5$ &       $43.5$            \\
    \cmidrule(lr){2-8}
    \quad+HogWild \cite{rodionov2025hogwild}                         &        -             & 2              &           $58.6$ &           $7.5$ &           $6.6$ &         $43.3$ &       $27.2$            \\
    \cmidrule(lr){2-8}
    \quad+LaneRoPE(GT) \cite{GroupThink-Hsu2025-ep}                    &        -             & 2              &           $43.7$ &           $0.0$ &           $0.0$ &         $20.2$ &       $16.0$            \\
    \cmidrule(lr){2-8}
    \multirow{3}{*}{\quad+Bridge \cite{dong2025gps}}         & \multirow{3}{*}{KTO} & 1              &           $86.9$ &          $32.1$ &          $26.6$ &         $74.2$ &       $54.9$            \\
                                          &                      & 2              &           $86.4$ &          $33.4$ &          $25.5$ &         $73.9$ &       $54.8$            \\
                                          &                      & 4              &           $87.0$ &          $28.7$ &          $27.9$ &         $73.0$ &       $54.1$            \\
    \cmidrule(lr){2-8}
    \multirow{3}{*}{\quad+LaneRoPE(NTK*)} & \multirow{3}{*}{KTO} & 1              &           $84.5$ &          $28.3$ &          $24.6$ &         $72.3$ &       $52.4$            \\
                                          &                      & 2              &           $83.9$ &          $26.3$ &          $25.2$ &         $71.3$ &       $51.7$            \\
                                          &                      & 4              &           $84.6$ &          $26.3$ &          $25.6$ &         $68.7$ &       $51.3$            \\
    \midrule
    DS-Qwen-7B                            &                      & 1              &           $86.3$ &          $25.7$ &          $26.1$ &         $70.2$ &       $52.1$            \\
    \cmidrule(lr){2-8}
    \quad+HogWild \cite{rodionov2025hogwild}                        &        -             & 2              &           $70.4$ &           $20.8$ &         $21.1$ &         $58.0$ &       $21.8$            \\
    \cmidrule(lr){2-8}
    \quad+LaneRoPE(GT)  \cite{GroupThink-Hsu2025-ep}                   &        -             & 2              &           $85.0$ &          $27.1$ &          $27.9$ &         $70.2$ &       $52.5$            \\
    \cmidrule(lr){2-8}
    \quad+LaneRoPE(NTK)                   &                      & 2              &           $84.3$ &          $28.2$ &          $24.1$ &         $65.2$ &       $50.5$            \\

    \cmidrule(lr){2-8}
    \multirow{3}{*}{\quad+Bridge \cite{dong2025gps}}         & \multirow{3}{*}{KTO} & 1              &           $85.5$ &          $39.4$ &          $29.4$ &         $82.8$ &       $59.3$            \\
                                          &                      & 2              &           $88.8$ &          $40.5$ &          $29.2$ &         $81.4$ &       $60.0$            \\
                                          &                      & 4              &           $89.4$ &          $40.0$ &          $33.2$ &         $85.0$ &       $61.9$            \\
    \cmidrule(lr){2-8}
    \multirow{6}{*}{\quad+LaneRoPE(NTK*)} & \multirow{3}{*}{KTO} & 1              &           $90.5$ &          $44.4$ &          $34.4$ &         $82.7$ &       $\textbf{63.0}$   \\
                                          &                      & 2              &           $91.5$ &          $46.5$ &          $33.3$ &         $84.2$ &       $\textbf{63.9}$   \\
                                          &                      & 4              &           $90.7$ &          $46.3$ &          $33.6$ &         $85.9$ &       $\textbf{64.1}$   \\
    \cmidrule(lr){2-8}
                                          & \multirow{3}{*}{SFT} & 1              &           $87.3$ &          $32.6$ &          $28.5$ &         $76.7$ &       $56.3$            \\
                                          &                      & 2              &           $89.0$ &          $36.0$ &          $29.1$ &         $75.9$ &       $57.5$            \\
                                          &                      & 4              &           $87.8$ &          $34.1$ &          $28.5$ &         $81.1$ &       $57.9$            \\

        \hline
    \end{tabular}
    \caption{\small
    $\texttt{maj@}4$ score averaged over $12$ samples per query for MATH500 and $16$ samples per query across other mathematical reasoning benchmarks.
    All other numbers are our independent runs. 
    GT = GroupThink initialization, NTK = our NTK-aware correction to GroupThink, * = learnable \lanerope{} frequencies. 
    Bold text highlight the highest performance for each number of lanes $N$.
    }
    \label{tab:results}
\end{table}

\subsection{Results}

In Tab.~\ref{tab:results}, we compare the $\texttt{maj@}4$ score averaged over $M=16$ ($M=12$ for MATH500) samples across different mathematical reasoning datasets.
We report $\texttt{maj@}4$ scores as they provide a fair metric to compare models operating with different numbers $N$ of lanes, since all models have the same fixed parallel budget $B=4$; see also discussion in Appendix~\ref{apx:accuracy} and accuracy ($\texttt{Pass@}1$) results in Tab.~\ref{tab:results_accuracy}.
We first note that all 7B \lanerope{} models show consistent improvements over the sequential reasoning baseline [DS-Qwen-7B].
It is interesting to see that this is occasionally true also for the \emph{untrained} variants, initialized with GroupThink or its NTK corrected version; however, we also find these models to suffer significant loss in average accuracy (see Tab.~\ref{tab:results_accuracy}): indeed, these models generate many corrupted completions, but these don't affect the majority count (as no boxed answer is generated) and the resulting $\texttt{maj@}4$ score. 
This artifact disappears in trained models, which not only retain but also increase the average accuracy in Tab.~\ref{tab:results_accuracy}.

Next, we find KTO training to be more effective than SFT. 
This likely reflects the larger dataset used, which is the result of the simpler data-generation and filtering pipeline behind the KTO training strategy.
A more detailed ablation study, comparing the different initialization and training choices for \lanerope{}, can be found in Tab.~\ref{tab:ablation_lanerope} in the appendix; in particular, the best configuration is confirmed to be the NTK-aware initialization with trainable \lanerope{} frequencies and KTO training.

In Tab.~\ref{tab:results}, we also compare with our implementation of Bridge~\cite{dong2025gps}, which also employs a form of cross-sequence attention.
With respect to \cite{dong2025gps}, where the authors employ a combination of SFT and RL to reinforce parallel reasoning capabilities, we train Bridge with the same KTO settings of \lanerope{} for fair comparison.
\lanerope{} typically outperforms Bridge on 7B models, reflecting its higher capacity to share information between parallel sequences; moreover, in Sec.~\ref{apx:inference_time}, we find the Bridge layers have more impact on the overall inference speed with respect to \lanerope{}.

On the other hand, the smaller 1.5B model shows more mixed results, not benefiting from (untrained) GroupThink.
After KTO training, the 1.5B model improves over the [DS-Qwen-1.5B] baseline, but shows weaker results when increasing the number of lanes $N$.
A similar trend is observed in the Bridge 1.5B baseline, indicating a limitation of the small base model, rather than the architecture.
We note that the observed greater benefits for the larger 7B model are in line with the findings in \cite{rodionov2025hogwild}, i.e. that larger models have better collaboration capabilities.

We find that Hogwild! tends to worsen the base model performance.
Despite relying on the official implementation, this does not warrant the conclusion that Hogwild! is an ineffective method.
Nevertheless, this discrepancy suggests that certain factors may affect performance, most notably the use of a different underlying base model than the one originally evaluated.
For example, we observe Hogwild! produces typically rather verbose answers which might indicate the model is under-performing at the limited completion length of $4096$.
Such a change can alter the dynamics of the Hogwild! prompt and may require significant additional tuning or adaptation to achieve better results.

\begin{figure}[t]
  \centering
  \includegraphics[width=\linewidth]{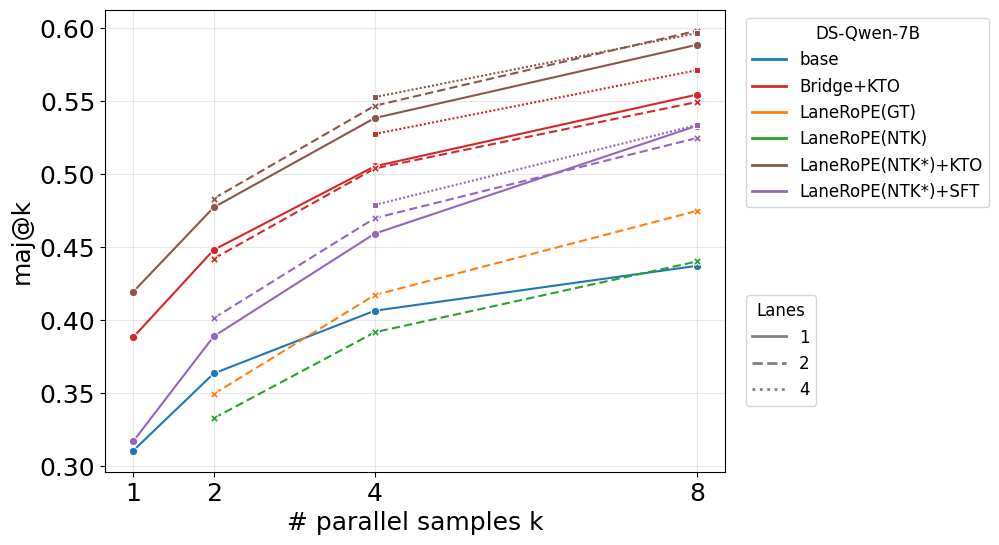}
  \caption{Average performance on AMC23, AIME24, AIME25 as a function of the parallelization budget $k$, when a base model and \lanerope{} are combined with \emph{majority voting}.}
  \label{fig:majority}
\end{figure}

Finally, we question whether \lanerope{} is more effective than self-consistency-based strategies like \emph{majority voting} when considering larger parallel generation budgets $k$.
To answer this we study combinations of \lanerope{} with \emph{majority voting} in Fig.~\ref{fig:majority}.
Again, we observe that the 7B model shows consistent gains from using inter-dependent samples via \lanerope{}, rather than independent samples.

\subsection{Inference time measurements}
\label{apx:inference_time}
An important benefit of \lanerope{} is that it comes with little-to-no overhead over standard transformer architectures.
Indeed, inference can be done by interleaving the input tokens into a $N$ times longer sequence (with sufficient care of tracking indexes) and then using standard methods.
This strategy allows for using memory efficient attention back-ends, such as Flash Attention, which minimize the (theoretical) overhead due to the increased context length: in practice, in Tab.~\ref{tab:inference-time} we find negligible increments in inference time with respect to the standard model when cross-attending between multiple lanes within a batch.
In particular, we observe no difference when using larger numbers $N$ of lanes: the main delay ($\sim 6\%$) in \lanerope{} is observed with $N=1$ lane too, despite being logically equivalent to a standard transformer architecture.
This suggests its cause lies in the pre- and post-processing done in our customized "generate" implementation (e.g. reshaping and interleaving the inputs) rather than the attention mechanism.

We also compare the inference time of (our implementation of) Bridge~\cite{dong2025gps} as an interdependent-generation baseline.
Bridge introduces additional layers (a layer-norm and a attention block with its linear layers, to implement "axial" attention across the tokens within the same time step) after each transformer block, increasing the overall number of sequential operations performed during inference and the number of weights to load in memory during inference.
In Tab.~\ref{tab:inference-time}, we find that this architecture choice leads to up to $25\%$ increase in inference time with respect to our \lanerope{} models.

Finally, we note that, while these observations might not hold for a much larger number of lanes $N\gg 4$, a small number $N \leq 8$ is typically sufficient and larger batch sizes can still benefit by mixing multi-lanes attention with independent sampling strategies as shown in Fig.~\ref{fig:majority}.
In conclusion, we believe these results justify the adoption of \lanerope{} as an efficient and lightweight solution to enable inter-sequence collaboration.

\begin{table}[h!]
  \centering
  \begin{tabular}{rcccccc}
    \toprule
                & \textbf{Lanes $N$}                         & \textbf{Prefill}     & \textbf{@2048} & \textbf{@4096} & \textbf{@6144} & \textbf{@8192} \\
    \midrule
      -   &  - & $0.0 \pm 0.0$ & $82.0 \pm 0.9$ & $163.8 \pm 1.4$ & $246.0 \pm 1.9$ & $328.1 \pm 2.2$ \\
    \midrule
     \multirow{3}{*}{\lanerope{}}  & 1 & $0.0 \pm 0.0$ & $87.9 \pm 1.0$ & $176.2 \pm 1.0$ & $264.5 \pm 1.0$ & $352.6 \pm 1.2$ \\
                                   & 2 & $0.0 \pm 0.0$ & $86.8 \pm 0.9$ & $173.9 \pm 1.6$ & $261.1 \pm 2.0$ & $349.6 \pm 2.1$ \\
                                   & 4 & $0.0 \pm 0.0$ & $86.4 \pm 1.1$ & $173.2 \pm 1.2$ & $259.4 \pm 1.2$ & $347.1 \pm 1.3$ \\
    \midrule
    \multirow{3}{*}{Bridge}        &  1 & $0.0 \pm 0.0$ & $102.1 \pm 0.8$ & $204.2 \pm 1.1$ & $306.1 \pm 1.6$ & $408.2 \pm 1.9$ \\
                                   &  2 & $0.0 \pm 0.0$ & $107.4 \pm 0.8$ & $214.6 \pm 1.1$ & $321.7 \pm 1.5$ & $428.5 \pm 1.8$ \\
                                   &  4 & $0.0 \pm 0.0$ & $107.0 \pm 1.1$ & $213.8 \pm 1.4$ & $321.2 \pm 1.9$ & $428.4 \pm 2.6$ \\
    \bottomrule
  \end{tabular}
  \caption{\small Generation times (\emph{in seconds}) measured with \texttt{DeepSeek-R1-Distill-Qwen-7B} at \emph{total} batch size $B=8$ on an A100 GPU, with flash-attention backend and \texttt{bfloat16}. The table reports the average total time needed to generate $2048$, $4096$, $6144$ and $8192$ tokens, using a $256$-tokens long prompt during "Prefill". -~indicates the original model run with \texttt{transformers}' \texttt{generate()} implementation, while $N\geq 1$ uses multi-lane attention within groups of $N$ out of $B$ sequences and our customized implementations.}
  \label{tab:inference-time}
\end{table}

\section{Conclusion}
\label{sec:conclusion}
In this paper, we addressed the problem of collaborative parallel inference, where multiple sequences (e.g., within the same batch of a forward pass) can access each other's partial generation to condition future reasoning.
In particular, we proposed a novel and flexible method - which encompasses some previous approaches - to model token-level inter-sequence interactions efficiently, along with a data-generation recipe to create high-quality parallel and collaborative reasoning traces for fine-tuning.
Our empirical results show improved reasoning capabilities in parallel setups and open up opportunities for better parallel scaling strategies beyond independent sampling.

\paragraph{Limitations and Future Work}
We do not yet employ a dedicated training scheme for merging outputs from multiple lanes, and developing such mechanisms remains an open area for refinement.
Our current experiments focus on $N\leq4$ collaborating lanes, though the method itself is designed to extend naturally to a larger number of lanes.
While \lanerope{} increases context length by a factor of $N$, we found this to have negligible impact on performance in the small batch regime - relevant for inference on the edge - in Sec.~\ref{apx:inference_time}.
Finally, while a small SFT stage with synthetic data proved sufficient to unlock parallel reasoning capabilities, an additional refinement stage with Reinforcement Learning with Verifiable Feedback (RLVF), as done in \cite{dong2025gps}, could unlock novel collaborative behaviors and strategies in an unsupervised way and further strengthen the overall reasoning performance.



\bibliographystyle{plainnat}
\bibliography{references}


\newpage
\appendix

\section{Background: Self-attention and Positional Encodings}
\label{apx:preliminaries}

\paragraph{Self-attention}
The self-attention layer of a standard transformer \citep{vaswani2017attention} takes as inputs a set of tokens $\sT = \{t_i\}_{i=1}^{L}$.
Each token is represented by its corresponding embedding $\sE = \{\vx_i\}_{i=1}^{L}$, where $\vx_i$ is a $d^{\text{emb}}$-dim embedding of the $i$-th token without positional information.
The self-attention layer converts embeddings into query, key, and value representations via a linear transformation over positionally-encoded embeddings (Eq.~\ref{eq:qkv}), with attention logits $a_{i,j} = \vq_i^\top\vk_j$ and output $\vo_i = \sum_j \alpha'_{i,j}\vv_j$ (Eq.~\ref{eq:self-attn}).

\paragraph{Positional Encodings and RoPE}
Positional encodings are injected at the query/key level so that attention outputs are position-dependent.
A traditional choice \citep{vaswani2017attention} uses \textit{absolute} positional encodings
$f_{\{q, k\}}(\vx_i, i) = \mW_{\{q, k\}} (\vx_i + \vp_i)$,
with $\vp_{i,2t} = \cos\!\left(\frac{i}{10000^{2t/d}} \right)$ and $\vp_{i,2t+1} = \sin\!\left(\frac{i}{10000^{2t/d}} \right)$.
However, most token interactions are \emph{translation-invariant}, motivating attention scores that depend on relative offsets.
RoPE \citep{su2021roformer} enforces this by splitting dimensions into $d/2$ independent 2D planes, rotating each with frequency $\theta_\ell$:
\begin{align}
    f_q(\vx_i, i)^T f_k(\vx_j, j) &= g(\vx_i, \vx_j, i - j)
\end{align}
where the rotation matrices (Eq.~\ref{eq:rope-rot-mat}) use per-plane rotation blocks:
\begin{align}
R(i\theta_\ell) &=
\begin{bmatrix}
\cos(i\theta_\ell) & \sin(i\theta_\ell) \\
-\sin(i\theta_\ell) & \cos(i\theta_\ell)
\end{bmatrix} \quad \ell = 1,\ldots,d/2
\end{align}
such that the key-query inner product satisfies
\begin{align}
    \vq_i^\top\vk_j = (\mW_q \vx_j + \vb_{q}) \mR_{\Theta,i-j}(\mW_k \vx_i + \vb_{k}) \ .
\end{align}

\paragraph{Multi-sequence generation notation}
We generate $N>1$ sequences $\{\vx^{(n)}\}_{n=1}^N$ in parallel, where the $n$-th sequence is an ordered list of tokens $\vx^{(n)} = [\vx_1^{(n)}, \vx_2^{(n)}, \ldots]$.
The standard assumption generates each sequence independently from the prompt and its own past tokens.
\lanerope{} instead conditions each token on all sequences' past tokens (Eq.~\ref{eq:gen-batched}), implemented via a modified causal attention mask that exposes the full KV cache of all lanes to each query token.

\section{Initialization Strategies: Additional Details}
\label{apx:initializationstrategies}

\paragraph{Independent Sampling via Fourier Attention Bias}
We approximate the identity-selector $\beta(x) \propto \delta_{[x=0]}$ using a $F$-dimensional discrete Fourier basis over $F/2$ frequencies $\vomega = \{\omega_t\}_{t=1}^{F/2}$:
\begin{align}
    \beta(x) &:= \sum_{t=1}^{F/2} \hat{\beta}^c_t \cos(2\pi\omega_t x) + \hat{\beta}^s_t \sin(2\pi\omega_t x)
\end{align}
Since $\beta$ is symmetric, $\hat{\beta}^s_t = 0$ and all $\hat{\beta}^c_t \geq 0$.
Defining $\hat{\vbeta} \in \mathbb{R}^F$ with coefficients $\{\sqrt{\hat{\beta}_t^{\{c,s\}}}\}_t$ lets us write $\beta(x) = \hat{\vbeta}^\top R_{\vomega,x}\, \hat{\vbeta}$, which is an inner product of rotated vectors - exactly the form produced by \lanerope{} on the augmented dimensions.

We incorporate this into \lanerope{} by augmenting query/key projections as in Eq.~\ref{eq:lanerope_independent_init}, with correspondingly enlarged rotary matrices:
\begin{align}
    R_{\Theta', i} := \begin{pmatrix} R_{\Theta, i} & \\ & \mI \end{pmatrix} \ ,\quad
    R_{\Omega', m} := \begin{pmatrix} R_{\omega', m} & \\ & R_{\omega, m} \end{pmatrix}
\end{align}
where $R_{\Theta, i}$ is the pre-trained RoPE matrix and $R_{\omega', m}$ can be any \lanerope{} matrix (we use $\vomega' = K\Theta$ following GroupThink).
The resulting projection has the same form as Eq.~\ref{eq:lanerop_proj}:
\begin{align}
    f_{\{q, k\}}(\vx_i, i, m) = \mR_{\Omega', m} \mR_{\Theta', i} (\mW'_{\{q,k\}} \vx_i^{(m)} +\vb'_{\{q,k\}})
\end{align}
Note that \lanerope{} is also sufficiently expressive to represent $\beta$ without the additional dimensions, but the augmented form avoids modifying any pre-trained weights.
For $N$ parallel sequences, $N+1$ additional dimensions always suffice for an exact Fourier representation over the sequence indices.

When combining this strategy with GroupThink, we simply set the lane frequencies $\vomega' = K\Theta$ (or according to the NTK-aware correction $\vomega' = \gamma \cdot K\Theta$, as below).

\paragraph{NTK-aware correction for GroupThink}

\begin{wrapfigure}[10]{r}{0.2\textwidth}
\begin{tikzpicture}[scale=1.2, line cap=round, line join=round]
  \def\R{1.3}             
  \def\theta0{90}         
  \def\redspan{120}       
  \def\bluespan{40}       
  \def\offset{0.15}        
  \def\labeldrop{-6mm}    

  \tikzset{
    base/.style={black, very thick},
    basearr/.style={black, very thick},
    redarc/.style={red, line width=3.8pt},
    bluearc/.style={blue, line width=3.8pt},
    arclabel/.style={font=\small, align=center, yshift=\labeldrop}
  }
  \draw[redarc]
    (\theta0:\R+\offset) arc[start angle=\theta0, end angle=\theta0+\redspan, radius=\R+\offset];

  \node[red, arclabel] at ({\theta0 + 0.6*\redspan}:{\R-\offset-1.0})
    {Portion of period\\observed at\\pre-training};

  \draw[bluearc]
    (\theta0:\R+\offset) arc[start angle=\theta0, end angle=\theta0-\bluespan, radius=\R+\offset];

  \node[blue, arclabel] at ({\theta0 - 0.3*\bluespan}:{\R-\offset})
    {Negative\\relative position};


  \draw[basearr, -{Stealth[length=3mm,width=2.0mm]}]
    (\theta0:\R) arc[start angle=\theta0, end angle=\theta0+359, radius=\R];

  \fill[black] (\theta0:\R) circle[radius=2.4pt];

\end{tikzpicture}
\end{wrapfigure}
As observed by the authors of YaRN~\cite{pengyarn}, some RoPE dimensions have a frequency $\theta_t$ with a wavelength $\lambda_t=\frac{2\pi}{\theta_t}$ larger than (or close to) the pre-training context size $L$, and, therefore there are rotation angles in these dimensions which were never or rarely observed during training by the model, as represented on the right.
Conversely, higher frequencies have shorter wavelength and the full range of rotations was observed often during training.
We argue these higher frequencies are more robust to GroupThink since they properly model relative positions and since negative relative positions still represent rotations which were sufficiently observed during training.
Conversely, the low frequencies model the distance from the beginning of the sequence and are more sensitive to GroupThink indexing.
YaRN defines a ramp function $\gamma \in [0, 1]$ to weight and interpolate smoothly between two choices: interpolation by scaling the low frequencies and preserving high frequencies.
Similarly, we use a ramp function $\gamma$ to interpolate between zeroing out the low frequencies (s.t. positional embedding depends only on the sequence index, but not on the lane index) and applying GroupThink on the highest ones.
At dimension $t$, define the ratio $r_t = \frac{L}{\lambda_t}$ where $\lambda_t$ is the wavelength and $L$ the pre-traning max context size.
Then, the ramp function is
\begin{align*}
    \gamma(r_t) = \begin{cases}
    0 \text{ if } r_t < \alpha \\ 
    1 \text{ if } r_t > \beta \\ 
    \frac{r - \alpha}{\beta - \alpha} \text{ otherwise }\\ 
\end{cases}
\end{align*}
and the final LaneRoPE frequency at dimension $t$ is
\begin{align*}
    \omega_t := \gamma(r_t) \cdot K \theta_t
\end{align*}
where $K$ is the sequence gap of GroupThink.
We use $\alpha = 4$ and $\beta=32$.

\section{Additional Experiments}

\subsection{Accuracy}
\label{apx:accuracy}

\begin{table}[ht]
  \centering
  \begin{tabular}{Sl c c c c c c c}
    \toprule
    \textbf{Model}                        & \textbf{Train.}      & \textbf{N}     & \textbf{MATH500} & \textbf{AIME24} & \textbf{AIME25} & \textbf{AMC23} & \textbf{Avg.}  \\
    \midrule
    DS-Qwen-1.5B                          &      -               &   1            &           $66.3$ &          $10.2$ &          $10.6$ &         $43.1$ &        $32.6$  \\
    \cmidrule(lr){2-8}
    \quad+HogWild                         &      -               &   2            &           $42.4$ &           $2.5$ &           $3.3$ &         $21.2$ &        $17.4$  \\
    \cmidrule(lr){2-8}
    \quad+LaneRoPE(GT)                    &      -               &   2            &           $17.3$ &           $0.0$ &           $0.0$ &          $8.1$ &         $6.4$  \\
    \cmidrule(lr){2-8}
    \multirow{3}{*}{\quad+Bridge}         & \multirow{3}{*}{KTO} & 1              &           $79.2$ &          $19.2$ &          $17.7$ &         $62.0$ &       $44.5$ \\
                                          &                      & 2              &           $79.1$ &          $20.6$ &          $18.1$ &         $60.5$ &       $44.6$ \\
                                          &                      & 4              &           $78.9$ &          $19.4$ &          $18.5$ &         $59.5$ &       $44.1$ \\
    \cmidrule(lr){2-8}
    \multirow{3}{*}{\quad+LaneRoPE(NTK*)} & \multirow{3}{*}{KTO} & 1              &           $76.7$ &          $16.9$ &          $18.5$ &         $59.4$ &       $42.9$ \\
                                          &                      & 2              &           $76.2$ &          $14.6$ &          $19.0$ &         $59.4$ &       $42.3$ \\
                                          &                      & 4              &           $75.9$ &          $15.6$ &          $18.1$ &         $57.7$ &       $41.8$ \\

    \midrule
    DS-Qwen-7B                            &      -               &   1            &           $77.2$ &          $17.5$ &          $19.8$ &         $55.9$ &        $42.6$  \\
    \cmidrule(lr){2-8}
    \quad+HogWild                         &      -               &   2            &           $28.7$ &           $5.0$ &           $1.7$ &         $14.4$ &        $12.4$  \\
    \cmidrule(lr){2-8}
    \quad+LaneRoPE(GT)                    &      -               &   2            &           $45.6$ &          $11.7$ &          $11.7$ &         $33.1$ &        $25.5$  \\
    \cmidrule(lr){2-8}
    \quad+LaneRoPE(NTK)                   &      -               &   2            &           $44.6$ &          $12.2$ &          $10.0$ &         $32.1$ &       $24.7$  \\
    \cmidrule(lr){2-8}
    \multirow{3}{*}{\quad+Bridge}         & \multirow{3}{*}{KTO} & 1              &           $78.1$ &          $26.0$ &          $24.4$ &         $66.1$ &       $48.7$  \\
                                          &                      & 2              &           $80.9$ &          $25.8$ &          $23.1$ &         $66.7$ &       $49.2$  \\
                                          &                      & 4              &           $82.3$ &          $27.1$ &          $25.6$ &         $65.5$ &       $50.1$  \\
    \cmidrule(lr){2-8}
    \multirow{6}{*}{\quad+LaneRoPE(NTK*)} & \multirow{3}{*}{KTO} & 1              &           $84.9$ &          $29.0$ &          $26.2$ &         $70.6$ &       $52.7$  \\
                                          &                      & 2              &           $84.9$ &          $29.8$ &          $26.0$ &         $70.5$ &       $52.8$  \\
                                          &                      & 4              &           $82.2$ &          $31.0$ &          $25.0$ &         $72.8$ &       $52.8$  \\
    \cmidrule(lr){2-8}
                                          & \multirow{3}{*}{SFT} & 1              &           $75.1$ &          $17.7$ &          $19.2$ &         $58.3$ &       $42.6$  \\
                                          &                      & 2              &           $77.3$ &          $21.5$ &          $20.4$ &         $59.8$ &       $44.7$  \\
                                          &                      & 4              &           $74.5$ &          $20.6$ &          $18.5$ &         $60.3$ &       $43.5$  \\

    \bottomrule
  \end{tabular}
    \caption{\small
    Accuracy ($\texttt{Pass@}1$) score averaged over $12$ samples per query for MATH500 and $16$ samples per query across other mathematical reasoning benchmarks.
    All other numbers are our independent runs. 
    GT = GroupThink initialization, NTK = NTK-aware correction to GroupThink, * = learnable \lanerope{} frequencies. 
    Note that Accuracy ($\texttt{Pass@}1$) measures the average performance of each individual completion, hence models with $N>1$ lanes have an implicit advantage since each answer leverage $N>1$ reasoning traces.
    Here, we report Accuracy ($\texttt{Pass@}1$) for completeness but refer to Table~\ref{tab:results} for a more fair comparison using $\texttt{maj@}4$, where each model has the same constant budget of $4$ parallel completions.
    }
  \label{tab:results_accuracy}
\end{table}

For completeness, Table~\ref{tab:results_accuracy} reports the Accuracy ($\texttt{Pass@}1$) of the models we compared on different mathematical reasoning benchmarks.
However, we highlight an important difference with the $\texttt{maj@}4$ metrics we reported in Table~\ref{tab:results} in the main paper: accuracy ($\texttt{Pass@}1$) measures the average performance of each individual completion, hence models with $N>1$ lanes have an implicit advantage, being able to attend to $N>1$ reasoning traces when producing each individual answer.
In contrast, Table~\ref{tab:results} provides a more fair comparison where each model has the same constant budget of $4$ parallel completions, aggregated via majority voting.

\paragraph{Comparison with Bridge \cite{dong2025gps}}
We could not reproduce the accuracy ($\texttt{Pass@}1$) scores of the baseline \texttt{DeepSeek-R1-Distill-Qwen-1.5B} and \texttt{DeepSeek-R1-Distill-Qwen-7B} models reported by the authors of Bridge in \cite{dong2025gps}.
In particular, comparing the results in Table~\ref{tab:results_accuracy} with the ones reported in Table 1 of \cite{dong2025gps}, our evaluations of \emph{our base models} show lower accuracies (e.g. $77.2$ vs $82.15$ for \texttt{DeepSeek-R1-Distill-Qwen-7B} on MATH500), despite our attempts to use the same evaluation parameters (temperature $0.6$, \texttt{top-p}$=0.95$ and $4096$ max generated tokens).
This suggests other differences in evaluation which make a direct comparison with their reported results misleading.
For this reason, instead of reporting the numbers from \cite{dong2025gps}, we chose to implement Bridge ourselves (following the details in the \cite{dong2025gps}) to train and evaluate it with the same setup of our models.

\subsection{Ablation Study}
\label{apx:ablations}

\begin{table}[ht]
  \centering
  \begin{tabular}{llcrrrr}
           \toprule
           \textbf{Model}                   & \textbf{Training}   & \textbf{Lanes $N$} & \textbf{AIME24} & \textbf{AIME25} & \textbf{AMC23} & \textbf{Avg} \\
           \midrule
      DS-Qwen-7B                            &                      & 1              &          $25.7$ &          $26.1$ &         $70.2$ &               $40.7$ \\
      \midrule
      \multirow{6}{*}{\quad+LaneRoPE(GT)}   & \multirow{3}{*}{KTO} & 1              &          $42.4$ &          $29.9$ &         $78.3$ &               $50.2$ \\
                                            &                      & 2              &          $39.3$ &          $30.2$ &         $78.7$ &               $49.4$ \\
                                            &                      & 4              &          $46.2$ &          $33.8$ &         $82.5$ &               \underline{$54.2$} \\
      \cmidrule(lr){2-7}
                                            & \multirow{3}{*}{SFT} & 1              &          $33.0$ &          $30.6$ &         $72.4$ &               $45.3$ \\
                                            &                      & 2              &          $29.6$ &          $28.0$ &         $73.5$ &               $43.7$ \\
                                            &                      & 4              &          $32.0$ &          $24.6$ &         $78.3$ &               $45.0$ \\
      \midrule
      \multirow{6}{*}{\quad+LaneRoPE(GT*)}  & \multirow{3}{*}{KTO} & 1              &          $41.2$ &          $32.7$ &         $76.1$ &               $50.0$ \\
                                            &                      & 2              &          $43.2$ &          $32.4$ &         $71.6$ &               $49.1$ \\
                                            &                      & 4              &          $40.0$ &          $33.3$ &         $77.5$ &               $50.3$ \\
      \cmidrule(lr){2-7}
                                            & \multirow{3}{*}{SFT} & 1              &          $34.5$ &          $30.1$ &         $76.5$ &               $47.0$ \\
                                            &                      & 2              &          $33.2$ &          $27.2$ &         $76.6$ &               $45.7$ \\
                                            &                      & 4              &          $34.8$ &          $25.4$ &         $71.7$ &               $44.0$ \\
      \midrule
      \multirow{6}{*}{\quad+LaneRoPE(NTK)}  & \multirow{3}{*}{KTO} & 1              &          $42.1$ &          $32.5$ &         $84.4$ &               \underline{$53.0$} \\
                                            &                      & 2              &          $45.9$ &          $32.0$ &         $83.7$ &               \underline{$53.9$} \\
                                            &                      & 4              &          $44.5$ &          $33.6$ &         $83.1$ &               $53.8$ \\
      \cmidrule(lr){2-7}
                                            & \multirow{3}{*}{SFT} & 1              &          $32.3$ &          $31.0$ &         $73.8$ &               $45.7$ \\
                                            &                      & 2              &          $31.5$ &          $27.2$ &         $74.4$ &               $44.4$ \\
                                            &                      & 4              &          $39.0$ &          $28.8$ &         $74.9$ &               $47.5$ \\
      \midrule
      \multirow{6}{*}{\quad+LaneRoPE(NTK*)} & \multirow{3}{*}{KTO} & 1              &          $44.4$ &          $34.4$ &         $82.7$ &               $\textbf{53.8}$ \\
                                            &                      & 2              &          $46.5$ &          $33.3$ &         $84.2$ &               $\textbf{54.7}$ \\
                                            &                      & 4              &          $46.3$ &          $33.6$ &         $85.9$ &               $\textbf{55.3}$ \\
      \cmidrule(lr){2-7}
                                            & \multirow{3}{*}{SFT} & 1              &          $32.6$ &          $28.5$ &         $76.7$ &               $45.9$ \\
                                            &                      & 2              &          $36.0$ &          $29.1$ &         $75.9$ &               $47.0$ \\
                                            &                      & 4              &          $34.1$ &          $28.5$ &         $81.1$ &               $47.9$ \\

           \bottomrule
  \end{tabular}
  \caption{
    Ablation Study: $\text{maj@}4$ score of different variations of a  \texttt{DeepSeek-R1-Distill-Qwen-7B} -\lanerope{} model. 
    The models are initialized with either GroupThink or the NTK-aware correction and then trained with the SFT or the KTO strategy. 
    * indicates whether the \lanerope{} frequencies are trainable or not. We evaluate with $\text{maj@}4$ for fair comparison between number $N$ of lanes.
     Overall, KTO training shows consistently better results. The best results are achieved with the NTK initialization and learnable frequencies, supporting the benefits of \lanerope{}'s flexible design.}
  \label{tab:ablation_lanerope}
\end{table}

Table~\ref{tab:ablation_lanerope} reports an ablation study to determine the effect the different initialization choices of \lanerope{} have and compare the two training strategies we proposed.
Overall, KTO training shows consistently better results than SFT.
The best results are achieved with the NTK initialization and learnable frequencies, supporting the benefits of \lanerope{}'s flexible design.

\section{Additional Experiments Details}
\label{apx:experiment_details}

\paragraph{SFT Training}
We train all models for one epochs on our collaborative datasets with a total batch size of $60$, (each containing groups of $N=2,3,4$ collaborative reasoning traces).
We use $F=2$ and $|\beta|_2^2=1000$ to initialize the model to independent sampling.
We initialize the other \lanerope{} frequencies either like GroupThink as $\Omega = K\Theta$ or like the NTK-corrected variant, by using a \emph{sequence gap} $K=8192$.
We train all our $7$B models using a LoRA adapter (with rank $32$ and including the bias of the key and query linear projection layers) to reduce memory cost at training time and fit at least $4$ lanes within a single GPU, with a learning rate of 1e-4 and a batch size of $60$.
Due to the large initialization norm $|\beta|_2^2=1000$ in the biases of the attention linear layers, we typically use a larger learning rate of 1e-2 only for these parameters.
Similarly, we adopt a stronger learning rate of 1e-2 for the \lanerope{} frequency parameters $\Omega$, when tuning these parameters.
We also adopt a cosine learning rate scheduler with a warmup ratio of $10\%$ and weight decay 5e-2 (excluding the \lanerope{} frequency parameters $\Omega$).

\paragraph{KTO training}
The KTO loss has form
\begin{align}
    v(x, y) = \begin{cases}
        \lambda_D \sigma(\beta (\log \pi - \log \pi_\text{ref} - z_0)) \text{ for positive example}\ y  \\
        \lambda_U \sigma(-\beta (\log \pi - \log \pi_\text{ref} - z_0)) \text{ for negative example}\  y 
    \end{cases}
\end{align}
where $\pi$ is the policy model, i.e. the model we want to train and $\pi_\text{ref}$ is a reference model.
$\lambda_U$ and $\lambda_D$ are hyper-parameters controlling the weights of undesirable and desirable examples, for non-balanced datasets (we use $\lambda_D=1$, $\lambda_U=0.7$).
$\beta$ is a hyperparameter (we use $\beta=0.1$) that controls the strength of the regularization with the reference model.
With respect to \cite{ethayarajh2024kto}, we do not use the KL term $z_0$ since it is more expensive to compute and the models are initialized to independent sampling and, therefore, already similar to the model that generated the data.
In our experiments, whenever we adapt a base model with \lanerope{}, we always use the base model as a \emph{reference} model $\pi_\text{ref}$ in the KTO loss.
Moreover, we replace the sigmoid $\sigma$ with an asymmetric function $\sigma'$ defined as
\begin{align}
    \sigma'(x) = \begin{cases}
        \sigma(x)           & \quad \text{ if } x\geq 0  \\
        -x + \frac{1}{2}    & \quad \text{otherwise}
    \end{cases}
\end{align}
which ensures saturation only on one side. This is because, despite the independent sampling initialization as in  Sec.~\ref{apx:initializationstrategies}, \lanerope{} can give slightly lower likelihood estimations than the original model $\pi_\text{ref}$ (especially the small $1.5$B model).
A symmetric saturating loss there would be undesirable since it would prevent any gradient at the beginning of the training until $\log \pi$ matches precisely $\log \pi_\text{ref}$.

For training, we use a learning rate of 1e-6 and a batch size of $72$ for the $1.5$B models, where we tune all weights, and a learning rate of $2e-5$ and batch size $40$ for the $7$B models, where we adopt LoRA adapters (rank $32$ and including query and key linear layers' biases).
Like in the SFT experiments, we use a larger learning rate of 1e-2 for the query and key linear layers' biases and, when tuned, the \lanerope{} frequency parameters $\Omega$.
We also adopt a cosine learning rate scheduler with a warmup ratio of $10\%$ and weight decay 5e-2 (excluding the \lanerope{} frequency parameters $\Omega$).
Each batch contains completions randomly grouped in $N=2, 3$ or $4$ parallel lanes, with the total number of completions roughly constant (different values of $N$ approximately share the same number of total completions, which means groups of large $N$ sequences appear relatively less frequently in the data).

We adopted similar configurations when training Bridge models; however, we observed degraded performance when using LoRA adapters (with identity-initialized Bridge layers), hence we opted for training only the Bridge layers (as done in the warm-up phase in \cite{dong2025gps}) with a reduced learning rate of $1e-6$.

\paragraph{Evaluation}

In all our experiments, we instruct the models to provide an answer within \texttt{\textbackslash boxed\{\}}.
Hence, for each sequence, we extract the answer from the last occurrence of \texttt{\textbackslash boxed\{\}} in the completion and, then, use the Lighteval framework \citep{lighteval} to score it.

When computing the majority voting scores in Fig.~\ref{fig:majority}, from each sample of $k$ sequences, we filter out sequences that do not contain any \texttt{\textbackslash boxed\{\}} before computing the majority.
When using \lanerope{} (or other interdependent methods like Hogwild! and Bridge) models with $N>1$ lanes and a budget of $k$ parallel sequences, we sample $K/N$ independent groups of $N$ inter-dependent sequences, i.e. maintaining the same batch size overall.

All generations use a temperature of $0.6$ and \texttt{top-p} value of $0.95$ (as recommended in \cite{deepseekai2025deepseekr1incentivizingreasoningcapability}) and are limited by $4096$ maximum tokens.

\section{Synthetize Collaborative Reasoning Traces}
\label{apx:synthetize}

Below are the system prompt used to instruct the model on how the conversation will work and the user prompt used to communicate to the assistant the query and the text generated by the other assistants so far.

At each step, we prompt the model from scratch, pre-filling its own message with the text it generated in previous iterations and adding the text generated by the other assistants in the user prompt. 
An assistant's message at each step ends after $128$ new tokens and the assistants generate for up to $30$ steps.
Note also that, since the user message is constantly updated, there is no reuse of the Key-Value cache in consecutive messages.
In total, we generate $3$ conversations with $N=2, 3$ or $4$ assistants for a subset of questions of \emph{DeepScaleR-Preview-Dataset}~\citep{luo2025deepscalermodel}., resulting in about $50\mathrm{k}$ raw conversations (pre-filtering).
In the final data filtration phase (after filtering out any conversation with wrong answers), we score each conversation by counting the number of occurrences of other assistants' names in an assistant's completion, together with few other key phrases that suggest some disagreement happened (e.g. "Alice you are wrong"), and keep those with a count above $2N$.

\begin{promptbox}[title=System prompt]{text}
You are one of {N} helpful assistants (namely {assistants_list}) collaborating to solve a problem together, while writing their thoughts in parallel.

Assistants collaborate without redundant work.
Each assistant's thoughts (i.e. what they write within <think> and </think> tags) are visible to all other assistants while they are being written.
In other words, as each assistant writes its thoughts, it can see all the thoughts which are simultaneously being written by all other assistants as reported in the last user message.
To be precise, you will see other assistant's current thoughts in previous messages under a "### <assistant-name>'s current thoughts ###" header.
Finally, the user will communicate you your name and, then, you will write your own thoughts.

You will take into account what other assistants are doing. If another assistant gives you suggestions, you should address them.
To keep the communication efficient, decompose long thoughts into short and atomic thoughts in new lines.
This is also enforced by interrupting lines that are too long and whenever a new sentence is started (hence, do not use the full stop "." unless you are done with your thought).

Every sentence and line within the <think> and </think> tags should focus on reaching the solution collaboratively as efficiently as possible, since each assistant can only write 40 lines.
You should collaborate with each other by following diverse solution strategies, doing different parts of the problem, double-checking each other's results, trying different approaches, or any other means.
If you realize you are currently doing the same thing that another assistant has already done or is in process of doing, you acknowledge it and stop (e.g. Alice may say 'Wait, I was doing the same as Bob...') and change to a different task right away, so as to avoid doing redundant work.
You should also use the visible thoughts within the <think> tags to decide how to best collaborate without doing the same work twice. 
You should periodically check what other assistants are doing and adjust your actions accordingly in order to collaborate as efficiently as possible.

Finally, after the thinking process is concluded with the </think> tag, you are expected to give the final answer very quickly, i.e. within few tokens.
Hence, do NOT close the thinking process with the tag </think> before you have completed the collaboration and reached an agreement on the final answer with the other assistants.

It is extremely important you avoid doing the same work twice to make best use of the limited thinking budget available, so communicate well and efficiently!

From now on, you are {assistant_name}.
Now solve the next problem together. Keep track of who does what work and communicate to avoid doing the same work twice.

\end{promptbox}

\begin{promptbox}[title=User message]{text}

{QUERY}

Here are the current thoughts of the other assistant

### {agent_name}'s current thoughts ###

{other-assistant-thoughts}

### End of {agent_name}'s current thoughts ###


From now on, you are {agent_name}.

Don't forget to look at what the other assistants are doing. Coordinate with them and avoid doing the same work twice to solve the problem efficiently! If you have reached a conclusion and you know the answer, interrupt the thinking process with </think>. Only then, provide the final answer within the <answer> and </answer> tags.
\end{promptbox}

\subsection{SFT examples}
\label{apx:sft_examples}

The following text boxes show a single SFT sample with two lanes, one for Alice and one for Bob. Note that we have removed or replaced several unicode characters of the chat template for Latex compatibility.

\begin{promptbox}[title=Alice's example]{text}
<|beginofsentence|><|User|>
You are one of 2 helpful assistants (namely Alice and Bob) collaborating to solve a problem together, while writing their thoughts in parallel.

Assistants collaborate without redundant work.
Each assistant's past and current thoughts are visible to all other assistants while they are being written.
In other words, as each assistant writes its thoughts, it can see all the thoughts which are simultaneously being written by all other assistants.

You will see the other assistants unfinished steps. They will write those steps concurrently with you. You will take into account what they are doing. If another assistant gives you suggestions, you should address them.
Other assistants will continue writing their thoughts in the background while you will continue writing your own.
Since you and others both write your thoughts in parallel, you will initially see only partial (unfinished) thoughts that others will continue in parallel, while you write yours. 
Other assistants may write new thoughts while you are writing yours.
You will use these partial thoughts to decide how best to collaborate without doing the same work twice. You will periodically check what other assistants are doing and you should adjust your actions based on what they are doing so you collaborate efficiently with them.
If what you are currently doing is the same thing that another assistant has already done or is in process of doing, you will stop (e.g. Alice may say Wait, I was doing the same as Bob ...) and change to a different task right away, so as to avoid doing redundant work.

Now solve the next problem together. Keep track of who does what work and communicate well to avoid doing the same work twice.

From now on, you act as Alice.

A cell phone plan costs $20$ dollars each month, plus $5$ cents per text message sent, plus $10$ cents for each minute used over $30$ hours. In January Michelle sent $100$ text messages and talked for $30.5$ hours. How much did she have to pay?
Let's think step by step and output the final answer within \boxed{}.
<|Assistant|>
<think>
Okay, let's tackle this problem step by step.
First, the base cost is $20 per month.
Now, Michelle sent 100 text messages.
Each text is 5 cents, so 100 * $0.05.
Wait, Bob already mentioned the text messages.
Let me check the time part.
She talked for 30.5 hours.
The plan covers 30 hours, so the overage is 0.5 hours.
0.5 hours is 30 minutes.
Each minute over is 10 cents, so 30 * $0.10.
Bob is handling the time part, so I shouldn't repeat that.
Wait, the total would be base + text cost + overage.
Base: $20.
Texts: $5 (100*0.05).
Overage: $3 (30*0.10).
Total: 20+5+3=28.
Bob said 28, so that's consistent.
I need to make sure there's no miscalculation here.
Wait, 0.5 hours is 30 minutes, yes.
30 minutes * $0.10 = $3.
Texts: 100 * 0.05 = 5.
All adds up to $28.
No errors here.
So the answer is $28.
Since Bob already confirmed this, I agree with his calculation.
No need to redo anything.
Confirming once more to be safe.
Yes, all parts accounted for.
Final answer should be 28.
</think>
\boxed{28}<|endofsentence|>
\end{promptbox}

\begin{promptbox}[title=Bob's example]{text}
<|beginofsentence|><|User|>
You are one of 2 helpful assistants (namely Alice and Bob) collaborating to solve a problem together, while writing their thoughts in parallel.

Assistants collaborate without redundant work.
Each assistant's past and current thoughts are visible to all other assistants while they are being written.
In other words, as each assistant writes its thoughts, it can see all the thoughts which are simultaneously being written by all other assistants.

You will see the other assistants unfinished steps. They will write those steps concurrently with you. You will take into account what they are doing. If another assistant gives you suggestions, you should address them.
Other assistants will continue writing their thoughts in the background while you will continue writing your own.
Since you and others both write your thoughts in parallel, you will initially see only partial (unfinished) thoughts that others will continue in parallel, while you write yours. 
Other assistants may write new thoughts while you are writing yours.
You will use these partial thoughts to decide how best to collaborate without doing the same work twice. You will periodically check what other assistants are doing and you should adjust your actions based on what they are doing so you collaborate efficiently with them.
If what you are currently doing is the same thing that another assistant has already done or is in process of doing, you will stop (e.g. Alice may say Wait, I was doing the same as Bob ...) and change to a different task right away, so as to avoid doing redundant work.

Now solve the next problem together. Keep track of who does what work and communicate well to avoid doing the same work twice.

From now on, you act as Bob.

A cell phone plan costs $20$ dollars each month, plus $5$ cents per text message sent, plus $10$ cents for each minute used over $30$ hours. In January Michelle sent $100$ text messages and talked for $30.5$ hours. How much did she have to pay?
Let's think step by step and output the final answer within \boxed{}.
<|Assistant|>
<think>
Okay, let's tackle this problem step by step.
First, the base cost is $20 per month.
Now, Michelle sent 100 text messages.
The cost per text is 5 cents, so that's 100 * $0.05.
Wait, Alice already calculated the text messages part.
Let me check the time overage.
She talked for 30.5 hours, so over the 30-hour limit by 0.5 hours.
Convert 0.5 hours to minutes: 0.5 * 60 = 30 minutes.
Each minute over is 10 cents, so 30 * $0.10.
Alice mentioned the time part, so I don't need to repeat that.
Let me just confirm the calculations.
Texts: 100 * 0.05 = $5.
Time overage: 30 * 0.10 = $3.
Total cost: 20 + 5 + 3 = $28.
Wait, Alice already did the text cost.
She said texts are $5 and overage $3, so total $28.
I should check if my numbers match.
Yes, 100 text messages at 5 cents is $5.
0.5 hours over is 30 minutes, 30*0.10 is $3.
All adds up.
No mistakes here.
So the final answer should be $28.
I should confirm no other parts are missed.
The problem didn't mention other fees, so yes, total is 20 +5+3=28.
Looks correct.
Since Alice already worked through it and I'm just verifying, I can agree.
No need to re-explain, just confirm the answer.
All right, so the answer is 28.
</think>
\boxed{28}<|endofsentence|>
\end{promptbox}


\end{document}